\pdfoutput=1
\PassOptionsToPackage{table}{xcolor}

\documentclass[11pt]{article}
\usepackage[]{acl}

\usepackage{times}
\usepackage{latexsym}
\usepackage{array} 
\usepackage{arydshln} 
\usepackage[T1]{fontenc}

\usepackage[utf8]{inputenc}

\usepackage{microtype}
\definecolor{Gray}{gray}{0.85}
\usepackage{subcaption}

\usepackage{latexsym}
\usepackage{diagbox}
\usepackage{graphicx}
\usepackage{hyperref}
\usepackage{subcaption}
\usepackage{algorithmic}
\usepackage[linesnumbered,ruled]{algorithm2e}

\usepackage{subcaption}

\usepackage{inconsolata}

\usepackage{lipsum}  
\usepackage{amsmath}
\usepackage{mathtools}
\usepackage{amsfonts}
\usepackage{kotex}
\usepackage{CJKutf8}

\usepackage[table]{xcolor}
\usepackage{tabularx}

\usepackage{adjustbox}
\usepackage{geometry}
\usepackage{booktabs}
\usepackage{amsmath}
\usepackage{amssymb}
\usepackage{graphicx}
\usepackage{multirow}
\usepackage{pifont}

\newcommand{\hwidth}[1]{%
  \noalign{\hrule \@height #1}%
}

\usepackage{cellspace}
\def\thickhline{\noalign{\hrule height.8pt}}

\title{Accelerating Multilingual Language Model for \\ Excessively Tokenized Languages}

\author{
        Jimin Hong \quad
        Gibbeum Lee \quad
        Jaewoong Cho \\
        KRAFTON \\
        \texttt{\{jimmy.h, pirensisco, jwcho\}@krafton.com}  \\
}

\begin{document}
\maketitle
\begin{abstract}
Recent advancements in large language models (LLMs) have remarkably enhanced performances on a variety of tasks in multiple languages. 
However, tokenizers in LLMs trained primarily on English-centric corpora often overly fragment a text into character or Unicode-level tokens in non-Roman alphabetic languages, leading to inefficient text generation.
We introduce a simple yet effective framework to accelerate text generation in such languages. 
Our approach involves employing a new language model head with a vocabulary set tailored to a specific target language for a pre-trained LLM. This is followed by fine-tuning the new head while incorporating a verification step to ensure the model's performance is preserved.
We show that this targeted fine-tuning, while freezing other model parameters, effectively reduces token fragmentation for the target language. 
Our extensive experiments demonstrate that the proposed framework increases the generation speed by a factor of 1.7 while maintaining the performance of pre-trained multilingual models on target monolingual tasks. 
\end{abstract}

\section{Introduction}\label{sec:intro}Modern large language models (LLMs)~\citep{gpt4openai,touvron2023llama, antropic2023claude} have exhibited remarkable capabilities for a variety of tasks in multiple languages~\citep{eloundou2023gpts,solaiman2023evaluating}. 
Although these models are predominantly trained on English-centric data, they have shown a significant degree of multilingual proficiency~\citep{bandarkar2023belebele}. 

However, when applied to non-alphabetic languages, these models often suffer from slower text generation due to English-centric tokenization ~\cite{rust-etal-2021-good, ahia2023all,petrov2023language}. 
Current tokenization techniques used in Large Language Models (LLMs) are data-driven and optimize segmentation based on the frequency of characters or bytes within a specific corpus~\citep{sennrich2015neural,kudo-2018-subword}. As a result, the tokenizers of multilingual models, which are heavily influenced by English-dominant training data, are predominantly composed of English subwords.
This leads to \textit{excessive fragmentation}, where non-English words are overly segmented into a large number of subword units~\cite{rust-etal-2021-good,ahia2023all,petrov2023language}. 
The autoregressive nature of LLMs further amplifies this inefficiency, as it sequentially requires the generation of text.

\begin{figure}[t]
\begin{adjustbox}{width=0.48\textwidth}
\centerline{\includegraphics[width=0.49\textwidth]{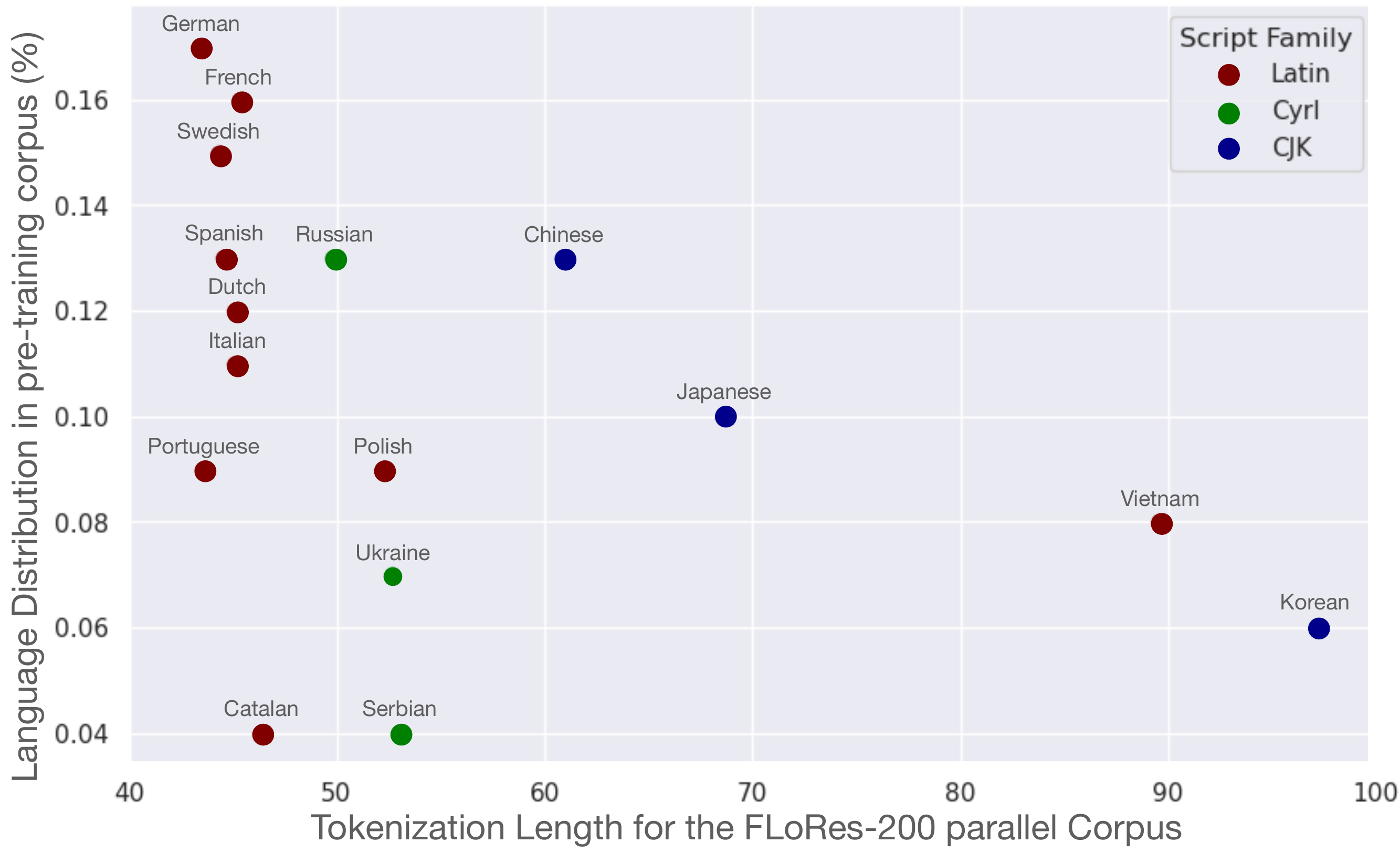}}
\end{adjustbox}
\caption{
\textbf{Analysis of tokenization lengths and language distribution in pretraining corpus with percentage >=0.04\%}
English script comprises 89.7\% of the corpus and has an average token length of 29.6 in FLoRes-200. 
The languages using the Chinese, Japanese, and Korean (CJK) scripts have longer tokenization lengths compared to those using Latin and Cyrillic scripts. 
Our primary focus is on languages that are excessively tokenized by English-centric tokenizers.
}
\label{fig:tokenization}
\vspace{-0.6cm}
\end{figure}

\begin{figure*}
\centerline{\includegraphics[width=0.85\textwidth]{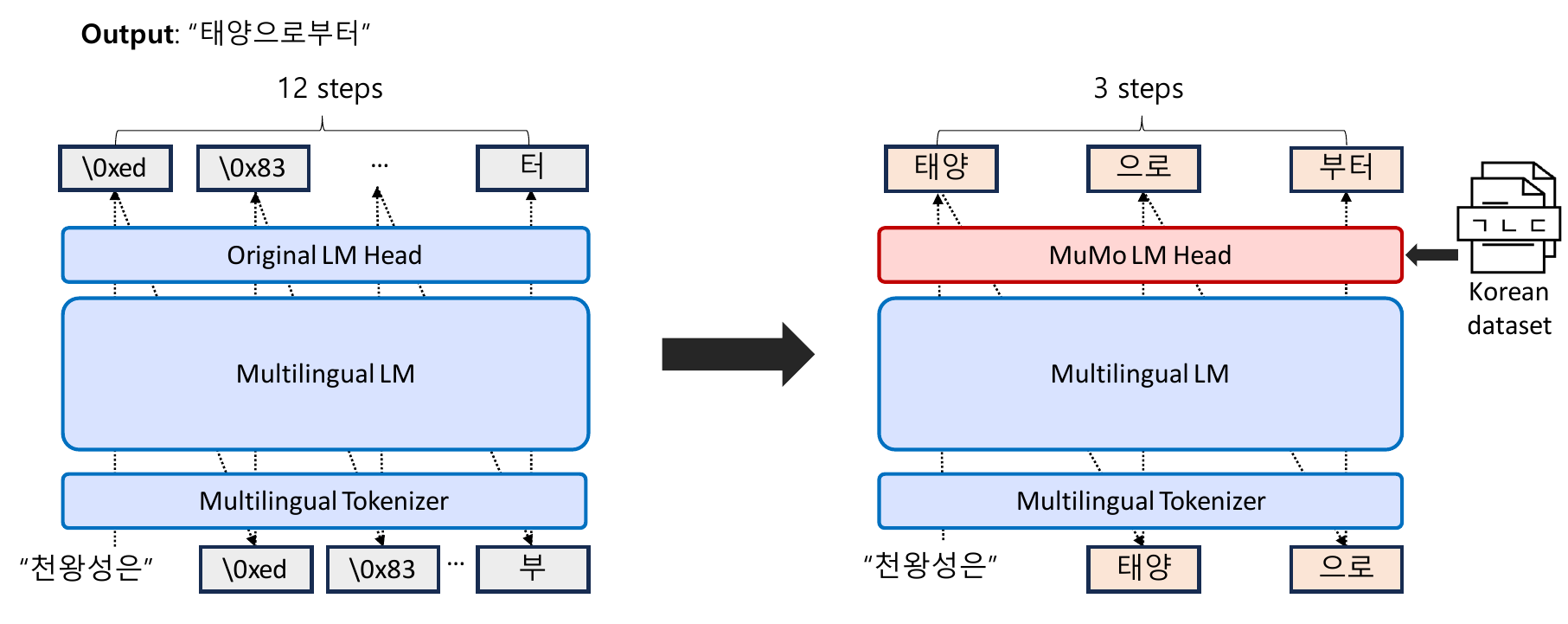}}
\caption{
\textbf{Overview of the proposed framework.} 
Illustration of (Left) the generation with a pre-trained multilingual model and (Right) the generation of MuMo Framework.
Given the Korean prefix ``천왕성은'' (\textit{Uranus is}), the model generates the consecutive phrase ``태양으로부터''(\textit{from the Sun}) that consisted of 3 morphemes (``태양'', ``으로'', ``부터'') in Korean. 
The generation with the pre-trained multilingual model faces inefficiency due to excessive fragmentation, requiring 12 steps to generate only 3 Korean morphemes.
However, the MuMo framework empowers the multilingual language model to generate multiple tokens in a single iteration by extracting a word from the Korean Vocabulary, requiring 3 steps.
}
\vspace{-0.5cm}
\label{fig:method}
\end{figure*}

To address these challenges, previous studies~\cite{wang2019improving,rust-etal-2021-good,cui2023efficient} have proposed replacing or augmenting the existing vocabulary of pre-trained multilingual models with language-specific vocabularies to more effectively encode monolingual text corpora. Specifically, \citet{rust-etal-2021-good} improved mBERT~\cite{devlin2019bert} by replacing its tokenizer with a monolingual one and incorporating an additional 100,000 pre-training steps. On the other hand, \citet{cui2023efficient} enhanced Llama~\citep{touvron2023llama} by expanding the Chinese vocabulary and further pre-training it on a 120GB text corpus that includes Chinese texts. However, this approach requires an extensive pre-training phase with a substantial amount of data.

Another approach to address the challenges is the use of small draft models~\citep{leviathan2023fast,chen2023accelerating}. 
These models generate draft output tokens, which are then verified by the original language model. 
However, a significant challenge arises when trying to identify or train a suitable small model that can handle multiple languages with reliable performance~\citep{conneau2019unsupervised,bandarkar2023belebele}.

In response to these challenges, our research introduces \textbf{MuMo}, accelerating \textbf{Mu}ltilingual language models for a targeted \textbf{Mo}nolingual text generation, particularly in non-alphabetic languages. MuMo incorporates a new vocabulary of a target language into the output layer, also known as the Language Model (LM) head, and predicts the next token from this expanded vocabulary. This approach requires training only the extended portion of the output layer and specific layers of the feed-forward network. Importantly, MuMo eliminates the need for extensive text corpora or a draft model, requiring only a modest corpus of the target language, approximately 44M tokens in size.
Empirical results across summarization, and translation tasks in Korean and Japanese demonstrate that MuMo significantly accelerates text generation, achieving over a 1.7-fold increase in speed without significantly compromising output quality.





\section{Related Work}\label{sec:related_work}

\begin{table}[t]
\begin{adjustbox}{width=0.48\textwidth}
\small
\begin{tabular}{cll}
\thickhline
\textbf{Lang}                                              & \multicolumn{1}{c}{\textbf{Word}} & \multicolumn{1}{c}{\textbf{Multilingual Tokens}} 
\\ \hline
\multicolumn{1}{c}{\multirow{1}{*}{\textsc{Ko}}}                        & 서울     & (``서'', ``\symbol{92}0xec'', ``\symbol{92}0xb8'', ``\symbol{92}0x9a'') 
\\ 
\hline
\multicolumn{1}{c}{\multirow{1}{*}{\begin{tabular}[c]
{@{}c@{}}\textsc{Ja}\end{tabular}}} & 
\begin{CJK*}{UTF8}{min}発売\end{CJK*}
& (\begin{CJK*}{UTF8}{min}``発''\end{CJK*}, ``\symbol{92}0xe5'',  ``\symbol{92}0xa3'', ``\symbol{92}0xb2'') \\
\thickhline

\end{tabular}
\end{adjustbox}
\caption{\textbf{Examples of the tokenization results.} 
These examples are preprocessed by the Llama tokenizer~\citep{touvron2023llama2}.
The target monolingual word are excessively segmented into byte units, when a suitable match is not found in the multilingual vocabulary. 
}
\label{tab:overly_tokenization}
\vspace{-0.4cm}
\end{table}


\paragraph{Tokenization Disparity}
Subword tokenization, a common approach in LMs, is typically data-driven.
Most of pre-trained tokenizers, which are often trained on predominantly English corpora, frequently result in excessive fragmentation of non-English scripts~\citep{rust-etal-2021-good,zhang2022robust}.
\citet{ahia2023all,petrov2023language} have found significant tokenization disparities across languages in popular LLMs~\citep{xue2020mt5,xue2022byt5,Scao2022BLOOMA1,gpt4openai}.
Our work endeavors to address the slowdown in inference that arises due to tokenization disparity in non-alphabetic languages.
\paragraph{Modifying Pre-trained Vocabulary}
Previous works have explored the adaptation of pre-trained vocabularies or the addition of new tokens~\citep{artetxe2019cross,rust-etal-2021-good,hong2021avocado,liu2023chipnemo}, these methods often necessitate extensive pre-training to integrate the new tokens effectively~\citep{wang2019improving,chau2020parsing,cui2023efficient,liu2023chipnemo}. 
In contrast, our MuMo framework sidesteps the need for fine-tuning the parameters of pre-trained models to preserve the original capabilities of the pre-trained language model. 
Efforts to select items of pre-trained embedding matrix have been made~\citep{abdaoui2020load,domhan2022devil,ushio2023efficient}, but these have not yielded significant speed up where the size of the embedding layer is relatively small ~\citep{bogoychev2023large}.
\paragraph{Accelerating LLM Inference}
The quest to accelerate inference in auto-regressive large language models (LLMs) has led to a variety of approaches. 
There has been a proliferation of systems specifically engineered for LLM inference~\citep{yu2022orca,sheng2023flexgen,xiao2023efficient}.
Our proposed methodology can be harmonically integrated with the aforementioned techniques.
Speculative decoding~\citep{leviathan2023fast,chen2023accelerating} have also been explored to increase inference velocity. 
However, the approach often relies on the assumption that a small model can maintain high fidelity when generating a series of multiple tokens. 
Moreover, acquiring a small yet competitive model may be tricky, especially in a multilingual setup~\citep{conneau2019unsupervised, bandarkar2023belebele}.
Our work distinguishes itself by specifically solving the inference inefficiency that arises from excessive fragmentation in the non-alphabetic context.
\paragraph{Parameter Efficient Cross-lingual Transfer Learning}
The \textit{curse of multilinguality}, which refers a trade-off between the language coverage and model capacity~\citep{conneau2019unsupervised}, is a significant issue even in massively multilingual models, such as mBERT, XLM-R, and mT5~\citep{devlin2019bert,conneau2019unsupervised,xue2020mt5,ansell2021madg}.
The problem has been mitigated through modular parameter-efficient adaptations of the multilingual models through lightweight adapters~\citep{houlsby2019parameter}: additional trainable parameters inserted into the transformer layers of model~\citep{pfeiffer2020mad,ustun2020udapter,vidoni2020orthogonal,parovic2022bad} for a target language.
These techniques bear a resemblance to ours, in that they involve training partial parameters of a language model with a small amount of target language corpus.
However, our goal is fundamentally different: we aim to accelerate the inference, whereas previous studies focus on improving the representational capability in target languages for multilingual models.

\section{Proposed Framework}\label{sec:method}

\begin{figure*}
\centerline{\includegraphics[width=0.85\textwidth]{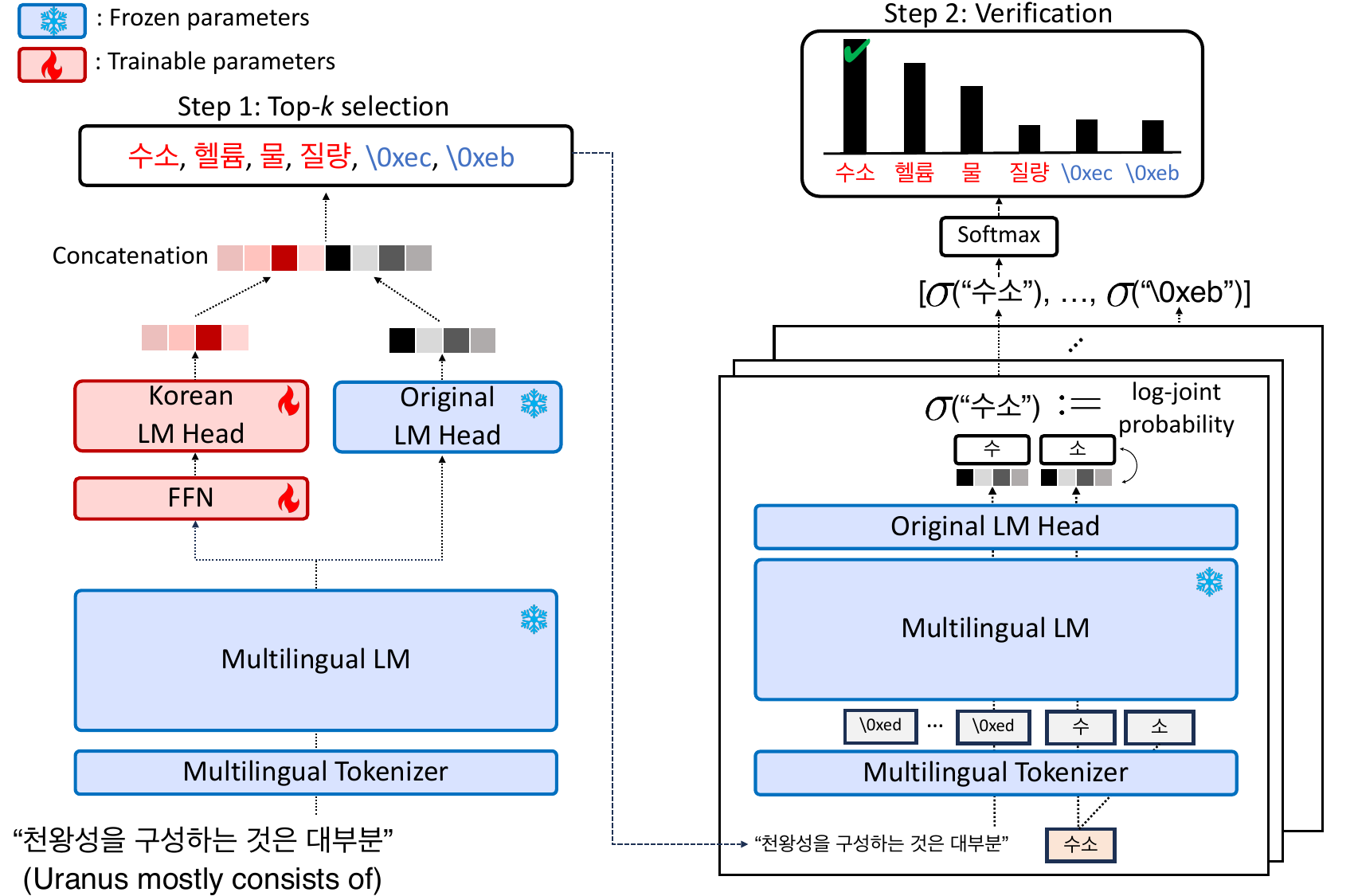}}
\caption{
\textbf{Illustration of a single-step prediction with MuMo.}
Initially, the MuMo LM Head $f_{\text{mumo}}$ selects the top 6 candidates.
Then, the pre-trained multilingual model verifies the feasibility of the candidates.
Among the modules in MuMo, the Target Monolingual LM head (the Korean LM Head in the figure) is only trained. 
}
\vspace{-0.4cm}
\label{fig:method}
\end{figure*}

We propose a framework named \textbf{MuMo} to accelerate the inference speed of a pre-trained multilingual LM for a non-alphabetic monolingual language via a given small monolingual dataset.
In the section, we introduce 1) the model architecture, 2) the fine-tuning process on a small targeted language dataset, and 3) the inference process of the proposed framework. 

\subsection{Model Architecture}\label{sec:model architecture}
We illustrate the model architecture of MuMo in Fig.~\ref{fig:method}. 
\paragraph{Pre-trained Multilingual Model}
We consider a setting in which a pre-trained multilingual model $f_{\text{multi}}$ is given. The model consists of 1) Transformer layers that consist of attention and feed-forward network, and 2) an output embedding layer called language model (LM) head.
We denote $\mathcal{V}_{\text{multi}}$ as the multilingual vocabulary set of the model objective, as $\mathcal{L}_{\text{MLE}}(p_{\text{multi}}, \mathbf{x}) = \sum_{t=1}^{|\mathbf{x}|}\log p_{\text{multi}}(x_t|\mathbf{x}_{<t})$, 


\paragraph{Target Monolingual LM Head}
The primary concept involves modifying pre-trained representations to predict a singular token unit within a target monolingual vocabulary $\mathcal{V}_{\text{mono}}$. 
The Target Monolingual LM head $f_{\text{mono}}$ projects the hidden representation $h$, which is composed of two main components: a feed-forward network (FFN) and an output linear layer, represented as $g_{\text{mono}}:  \mathbb{R}^{d_{_{\text{mono}}}} \rightarrow\mathbb{R}^{|\mathcal{V}_{\text{mono}}|}$:
\begin{equation} \label{eq1}
\text{FFN}(h) =  q(W_1^{\top}h)W_2 \in 
\mathbb{R}^{d_{\text{mono}}},
\end{equation}
where $W_1 \in \mathbb{R}^{d_{\text{multi}}\times d_{\text{ffn}}}$ and $W_2 \in \mathbb{R}^{d_{\text{ffn}}\times d_{\text{mono}}}$ are the weight matrices, $q$ is non-linearity function, 
and $d_{\text{mono}}$ represents the dimension of the target language representaiton. 
We set $d_{\text{ffn}}$ as $d_{\text{multi}}/4$, and the non-linearity function $q$ as SwiGLU~\citep{shazeer2020glu}.
The output linear layer $g_{\text{mono}}$ then generates a subword token:
\begin{equation} \label{eq1}
\begin{split}
f_{\text{mono}}(h) =  g_{\text{mono}}(\text{FFN}(h)) \in 
\mathbb{R}^{|\mathcal{V}_{\text{mono}}|}.
\end{split}
\end{equation}

\paragraph{MuMo LM Head}
Note that the output space of $f_{\text{mono}}$ is restricted to tokens in the $\mathcal{V}_\text{mono}$.
Inspired by \citet{lan2023copy}, we simply extend the $f_{\text{mono}}$ by concatenating the output linear layer of pre-trained multilingual model.
This is particularly useful when there is no suitable token in $\mathcal{V}_\text{mono}$ to predict, such as special symbols or alphabet-based tokens for non-alphabet languages.

Formally, given context representation $h_{t-1}$, the output of the MuMo LM head is computed as:
\begin{gather}
    f_{\text{mumo}}(h_{t-1}) = \nonumber \\
    [f_{\text{multi}}(h_{t-1}) ; f_{\text{mono}}(h_{t-1})] \in \mathbb{R}^{|\mathcal{V}_{\text{multi}}|+|\mathcal{V}_{\text{mono}}|} 
\end{gather}
where the symbol $;$ indicates the concatenation of two vectors,
and the $f_{\text{mumo}}$ indicates the output of the MuMo LM head.
Thus, the MuMo LM head is composed of a combination of the pre-trained language model head and Target Monolingual LM head.

\subsection{Fine-tuning}
In the proposed framework, we only fine-tune the target monolingual LM head $f_{\text{mono}}$ leveraging a small given target monolingual dataset. 
Note that the parameters of the pre-trained multilingual model remain frozen during the process. The model is fine-tuned by maximizing the log-likelihood of a sequence:
\begin{equation}
\resizebox{0.43\textwidth}{!}{
$\underset{{f_{\text{mono}}}}{\max}~
\mathcal{L}_{\text{MLE}}(p_{\text{mumo}},\mathbf{x}) = \sum_{t=1}^{T}\log p_{\text{mumo}}(x_t|\mathbf{x}_{<t})$
},
\end{equation}
where $p_{\text{mumo}}(x_t|\mathbf{x}_{<t})= \operatorname{Softmax}(f_{\text{mumo}}(h_{t-1}))$.

\subsection{Inference}
\label{sec:verify}

Despite the availability of direct generation based on the $p_{\text{mumo}}$, the newly initialized Target Monolingual LM head, which is trained on limited data, may be constrained by generalization capabilities beyond the training dataset.
The key concept is to leverage the probabilistic knowledge acquired by the pre-trained model $p_{\text{multi}}$, which has been extensively trained on large text corpora.

\subsubsection{Step 1: Top-$k$ Selection}

Initially, we select top-$k$ candidates based on the probability $p_{\text{mumo}}(x_t|\mathbf{x}_{<t})$. 
We set $k$ as 10 for all experiments.
Given the fact that we do not modify the input embedding of the pre-trained model, we are unable to feed the predicted word if a word does not belong in $\mathcal{V}_{\text{multi}}$ during the subsequent iteration.
Instead, we input the predicted word as the tokenization units of the pre-trained vocabulary.
For example, let's consider the Korean word ``수소'', which corresponds to a sequence of two tokens (``수'', ``소'') in $\mathcal{V}_{\text{multi}}$. 
If the Korean word ``수소'' is selected among the Top-$k$ candidates, we employ these two multilingual tokens.

\subsubsection{Step 2: Verification}
\label{sec:verify}
Then, the \textit{feasibility} of these potential completions is measured using the log-joint probability distribution over $p_{\text{multi}}$.
To account for shorter sequences naturally having higher scores~\citep{jean2015montreal,murray2018correcting}, we normalize each candidate's score by its token length.

We measure the \textit{feasiblity} for a candidate sequence as follows:
\begin{equation} \label{eq3}
\begin{split}
\sigma(\mathbf{c}^i) & = \frac{1}{l^i} \sum_{k=1}^{l^i}{\log p_{\text{multi}}(c_{t+k}^i|c_{<t+k}^i,\mathbf{x}_{<t})},
\end{split}
\end{equation}
where $\mathbf{c}^i$ symbolizes a predicted token within the top-$k$ candidates, $p_{\text{multi}}$ represents the probability as determined by the pre-trained multilingual model, and $l^{i}$ corresponds to the sequence length of the candidate $\mathbf{c}^i$.

From the $k$ candidates, the ultimate prediction can be derived from both deterministic and stochastic manners, depending on decoding strategies.



  
              
  

\section{Experiments}\label{sec:experiments}\subsection{Setup}


\paragraph{Languages}
As a case study,
we focus on two non-roman alphabetical languages: Korean and Japanese. 
Since we aimed to utilize a pre-trained model with a reasonable level of effectiveness in the target language, it is essential that the language is explicitly mentioned as being trained within the pre-training corpus. 
In this context, we considered languages included in the Llama-2~\citep{touvron2023llama2} pre-training corpus.
Moreover, the chosen language needed to exhibit the excessive fragmentation problem~\citep{ahia2023all,petrov2023language} by the English-centric pre-trained tokenizer. (See the Figure~\ref{fig:tokenization})
This criterion led to the exclusion of most European languages such as French, German, and Portuguese.
Finally, we conduct a study on multiple tasks, necessitating the existence of an instruction dataset for the target language.
Due to these considerations, we only implement the experiment in Korean and Japanese.

\paragraph{Model}
We utilize the Llama-2 13B model~\citep{touvron2023llama2} for all experiments.
We observed some language alignment discrepancies between instructions and responses when using the Llama-2 13B chat model.\footnote{\href{https://huggingface.co/meta-llama/Llama-2-13b-chat}{meta-llama/Llama-2-13b-chat}}
To address the issue, we 
conduct multilingual instruction tuning~\citep{muennighoff2022crosslingual} for English, Korean, and Japanese languages using the ShareGPT and Alpaca~\citep{chen2023msift}.
This process improve the model's fluency in each language~\citep{muennighoff2022crosslingual, chen2023monolingual}.
We also report our results test on Llama-1 13B~\citep{touvron2023llama} in Appendix.
\paragraph{Implementation of MuMo}
To construct targeted monolingual vocabularies in MuMo Framework, we levergage the tokenizers from the off-the-shelf model, as shown in Table \ref{tab:tokenizer_source}. 
We selected monolingual tokens by filtering vocabulary items based on the Unicode range of each monolingual script.  
Additionally, we excluded items from the selection if they were already present in the pre-trained vocabulary.
In terms of the preprocessing algorithm, we employ a forward maximum matching strategy to identify words in a target language vocabulary.
This strategy identifies the longest sequence of tokens that aligns with a word in the target language vocabulary.


Regarding the initialization of $g_{\text{mono}}$, we utilize the LM head of the pre-trained multilingual model. 
For example, when the Korean word "태양" is tokenized into subword units (``\symbol{92}0xed'', ..., ``\symbol{92}0x91'') using the pre-trained vocabulary, we initialize the Korean LM head of "태양" by taking the mean of the corresponding subword embeddings of the multilingual LM head. 
This process ensures that the initialized embeddings of Target Monolingual head represent the original word in the multilingual context.
\begin{table}[t]
\centering
\begin{adjustbox}{width=0.47\textwidth}
\begin{tabular}{lll}
\toprule
\textbf{Language}   & \textbf{Language Family}        & \textbf{Pre-trained Tokenizer}\\
\hline
Korean & Koreanic & \href{https://huggingface.co/EleutherAI/polyglot-ko-12.8b}{EleutherAI/polyglot-ko-12.8b} \\
\hline
Japanese & Japonic &  \href{https://huggingface.co/rinna/japanese-gpt-neox-3.6b}{rinna/japanese-gpt-neox} \\
\bottomrule
\end{tabular}
\end{adjustbox}
\caption{
\textbf{Selected languages and tokenizers.
}
We utilize the tokenizers to construct $\mathcal{V}_{\text{mono}}$ in each language.
}
\label{tab:tokenizer_source}
\end{table}
\paragraph{Fine-tuning}

We only train the Target Monolingual LM head $g_{\text{mono}}$ with the translated ShareGPT and Alpaca datasets~\citep{chen2023msift} in Korean, and Japanese.
The training is done with 1500 steps with one batch consisting of 128 examples. 
We use the AdamW~\citep{loshchilov2018decoupled} optimizer with a learning rate of 0.001, weight decay of 0.01, and 150 steps of warm-up.

\begin{table*}[t]
\centering

\begin{adjustbox}{width=0.99\textwidth}
\begin{tabular}{c|c|cccc|ccc}
\thickhline
{}     & {}      & \multicolumn{4}{c|}{\textbf{Summarization (0-shot)}}                                                           & \multicolumn{3}{c}{\textbf{Translation (3-shot)}} 
\\ \hline
{\textbf{Lang}} & {\textbf{Method}} & 
\textbf{ROUGE-2} & \textbf{ROUGE-L} & \textbf{Tokens/sec} & \textbf{Speed Up} & 

\textbf{BLEU} & \textbf{Tokens/sec} & \textbf{Speed Up} 


\\ \hline
\multirow{5}{*}{\textsc{Ko}}        
& Vanilla Decoding                        & \textbf{20.7}        &  \underline{36.1}       &  28.9                           & 1.00x       & \underline{21.2}               & 29.8  
& 1.00x


\\

& Spec. (w/o Rejection)                        & 18.7        & 33.5       & \underline{35.2}                           & \underline{1.21x}       
& 18.6               & \underline{36.5} & \underline{1.22x}
\\

& Spec.                        & 20.3        &  35.2       &  27.5                           & 0.95x       & 21.5               & 29.2  & 0.98x 

\\
& Shortlisting  & \underline{20.5}        &  \textbf{36.3}      &  30.6                           & 1.06x       & 19.5               & 32.7  
& 1.03x

\\

& MuMo (Ours)                        & 20.3        & 35.9       &   \textbf{55.3}   &  \textbf{1.92x}    
&\textbf{21.7} & \textbf{50.9} &  \textbf{1.70x}
\\

\hline
\multirow{5}{*}{\textsc{Ja}}        
& Vanilla Decoding                         & 11.3        & \textbf{26.6}        &     29.3               & 1.00x       
& \textbf{26.3}              &  33.4
& 1.00x

\\ 

& Spec. (w/o Rejection)                        & 10.8        & 24.2       & \underline{35.4}                           & \underline{1.21x}       & 22.7 & \underline{39.9} & \underline{1.21x}
\\
& Spec.                        & \textbf{11.6}        &  \underline{26.5}       &  28.5                           & 0.97x       & \underline{26.0} & 29.7   & 1.03x   
\\
& Shortlisting      & 11.4        & 26.3        &     30.3               & 1.03x       
& 25.2              &  34.9
& 1.04x

\\
& MuMo (Ours)                        & \textbf{11.6}        & 26.3         &    \textbf{59.2}                         & \textbf{2.02x}      & 24.3              & \textbf{58.3} & \textbf{1.75x} 
\\






\hline

\end{tabular}
\end{adjustbox}
\caption{
\textbf{Comparative study of Language Model (LM) inference speed.} The column labeled ``\textbf{Speed Up}'' represents the relative performance improvement in inference speed compared to the vanilla decoding method. The highest performance in each category is highlighted in \textbf{Boldface}, and the second highest score is \underline{underlined}. 
All models use sampling-based decoding. 
MuMo outperforms the compared baselines in the inference speed.
Detailed information about the generation hyperparameters, including those used for sampling-based decoding, can be found in Appendix~\ref{appendix:generation}.
}
\vspace{-0.3cm}
\label{tab:main}

\end{table*}



\paragraph{Evaluation}
We choose two representative generation tasks: summarization and translation.
For summarization, we use 500 examples from XLSum~\citep{hasan2021xl}, and 
for translation, we use 500 examples from the FLoRes-200~\citep{goyal2022flores} dataset.
We translate English sentences to each target language sentence.

For each task, we report 0-shot results for summarization, and 3-shot results for translation. 
We set the maximum sequence length as 512.
We utilize flash-attention 2~\citep{dao2023flashattention} and bfloat16 types for text generation.

\paragraph{Metrics}
In the summarization task, we gauge the reliability of the generated content by calculating the \textbf{ROUGE-2} and \textbf{ROUGE-L}~\citep{lin2004rouge} scores, averaging the results across 5 different generated summaries. 
Likewise, for the translation task, we measure the quality of the translations by computing the \textbf{BLEU}~\citep{papineni2002bleu} score, again averaging over 5 translation results.\footnote{We utilize SacreBLEU scores with the signature BLEU |nrefs:1 |case:mixed |eff:no |tok:{ko,ja}-mecab|smooth:exp |version:2.2.0.}
We report \textbf{Tokens/sec} to measure the inference speed of the models.


\subsection{Baselines}
We consider the following several baselines for the comparison with the proposed method. Note that all the baselines are implemented instruction-tuned model with multilingual instruction dataset~\citep{chen2023msift}.

\paragraph{Vanilla Decoding}
The autoregressive generation is to sequentially sample the subsequent word based on the probability distribution over the pre-trained vocabulary. 
This approach serves as the standard against which improvements are measured.
Accounting for the nature of task, all the baselines and our framework utilizes sampling-based decoding strategy with temperature as 0.1, $k$ as 10 for top-$k$ sampling~\citep{fan2018hierarchical} and $p$ as 0.7 for nucleus sampling~\citep{holtzman2019curious}.

\paragraph{Speculative Decoding}
Speculative decoding approach~\citep{chen2023accelerating,leviathan2023fast} employs a preliminary "draft" model to rapidly generate a set of token candidates at each decoding step. 
Subsequently, these candidates undergo a validation process by the original language model to ascertain their likelihood as plausible continuations of the text. 
We implement two variants of this method: one with the capability to reject unsuitable candidates \textbf{(Spec.)} and another without its rejection module \textbf{(Spec. w/o Rejection)}.
For the draft model, we utilize Llama-2 7B~\citep{touvron2023llama2}.
Following the implementation of \citet{chen2023accelerating}, we generate 5 draft tokens at each iteration.

\paragraph{Lexical Shortlisting }
Lexical Shortlisting \textbf{(Shortlisting)}~\citep{abdaoui2020load,ushio2023efficient}, or vocabulary selection, is the approach that optimizes the decoding process by allowing it to generate a word within a set of tokens during the inference stage~\citep{ushio2023efficient}.
We implement to filter out tokens that are not present within the corresponding target language subset of the mC4 corpus~\citep{xue2020mt5}, as \citet{ushio2023efficient}.

\begin{table*}[t]
\centering

\begin{adjustbox}{width=0.99\textwidth}
\begin{tabular}{c|c|c|cccc|ccc}
\thickhline
{} & {} & {} & \multicolumn{4}{c|}{\textbf{Summarization (0-shot)}}                                                           & \multicolumn{3}{c}{\textbf{Translation (3-shot)}}

\\ \hline
\textbf{Method} 
& \textbf{Update Param.} 
& \textbf{Dataset size (Tokens)} &
\textbf{ROUGE-2} & \textbf{ROUGE-L} & \textbf{Morphemes/sec} & \textbf{Speed Up} & 

\textbf{BLEU} & \textbf{Morphemes/sec} & \textbf{Speed Up} 

\\ \hline

Vanilla Fine-tuning &13.0B  &  44M                 & \textbf{21.0}        &  36.0       &  9.8                           & 1.00x       & \underline{21.4}               & 10.1  
& 1.00x

\\

Vocabulary Expansion &13.1B & 44M       & 13.7        & 23.1       & \underline{17.1}                           & \underline{1.92x}   & 12.3               & \underline{20.2} & \underline{2.00x}
\\

Vocabulary Expansion$^\dagger$ &13.1B   & 60B + 44M        & 20.3        &  \textbf{37.3}       &  \textbf{20.5}                           & \textbf{2.12x}       & 20.3               & 23.1  & \textbf{2.29x }

\\

MuMo (Ours) & 70M &  44M         & \underline{20.5}        & \underline{36.3}       &   15.3   &  1.73x      

& \textbf{21.7} & 17.2 &  1.71x

\\ 

\hline

\end{tabular}
\end{adjustbox}
\caption{
\textbf{Comparsion with the fine-tuning strategies.}
The column labeled ``\textbf{Speed Up}'' represents the relative performance improvement in inference speed compared to Vanilla Fine-tuning.
Vocabulary Expansion$^\dagger$ was pre-trained on over 60B tokens, comprised of both Korean and English text corpora.
Other methods are only trained with the instruction dataset (44M tokens)~\citep{chen2023msift}, ShareGPT and Alpaca translated in Korean.
The \textbf{Boldface} signifies the superior performances, and the second highest score is \underline{underlined}.
}
\label{tab:vocab_expansion}
\end{table*}

\subsection{Results}

Table~\ref{tab:main} shows the generation results in both summarization and translation tasks.
For the summarization task in Korean, MuMo outperforms all baselines in terms of speed, achieving a 1.92x speed-up over the Vanilla Decoding while maintaining competitive ROUGE scores. 
In translation, MuMo again demonstrates superior efficiency with a 1.70x speed-up and even shows an improvement in BLEU score compared to Vanilla Decoding. 

In the case of Japanese, the results are similar, with MuMo achieving a 2.02x speed-up in summarization and a 1.75x speed-up in translation. 
The ROUGE and BLEU scores for MuMo are on par with or slightly below Vanilla Decoding, indicating that the increase in speed does not significantly compromise the quality of the output.

\textbf{Shortlisting} shows only marginal gains in speed across both languages and every tasks, while preserving the generation capability.
This is likely because the relative computational cost of processing the embedding matrix is reduced in larger models, making vocabulary reduction less impactful~\citep{berard2021efficient,ushio2023efficient}.
On the other hand, the \textbf{Spec.} heavily relies on the capacity of the draft model, as shown as the comparison with \textbf{(Spec. w/o Rejection)}.
If the draft model lacks of sufficient multilingual capacity, it may not generate high-quality candidates, leading to a lower acceptance rate by the original model and thus reduced efficiency.

The superior performance of MuMo in terms of inference speed can be primarily attributed to its capability to predict larger linguistic units compared to those in the pre-trained vocabulary. 
We found that the target language tokens in $\mathcal{V}_{\text{mono}}$ are typically tokenized into 3-4 separate tokens in $\mathcal{V}_{\text{multi}}$, suggesting that the decoding step could potentially be reduced by 3-4 times. 
It is hypothesized that the inference speed is significantly influenced by the disparity between the pre-trained multilingual vocabulary and the target language. 
\section{Further Analysis}\label{sec:analysis}


\begin{table*}[t]
\centering
\begin{adjustbox}{width=0.7\textwidth}
\begin{tabular}{cl|cc|c}
\thickhline
\multicolumn{2}{c|}{}                       & \multicolumn{2}{c|}{\textbf{Summarization (0-shot)}}                 & \textbf{Translation (3-shot)}   \\ \hline
\multicolumn{2}{c|}{\textsc{\textbf{LM Head Initialization}}} & \multicolumn{1}{c|}{\textbf{ROUGE-2}}       & \textbf{ROUGE-L}       & \textbf{BLEU}          \\ \hline
\multicolumn{2}{c|}{\textsc{Mono-init}}              & \multicolumn{1}{c|}{\textbf{20.7}} & 36.2          & 21.5          \\ \hline
\multicolumn{2}{c|}{\textsc{Random-init}}            & \multicolumn{1}{c|}{19.2}          & 35.5          & 17.2          \\ \hline
\multicolumn{2}{c|}{\textsc{Multi-Init}}             & \multicolumn{1}{c|}{20.3}          & \textbf{36.3} & \textbf{21.7} \\ 
\hline
\end{tabular}
\end{adjustbox}
\caption{\textbf{Comparative analysis for the initialization strategy.}
\textsc{Mono-Init} signifies to leverage the pre-existing embedding representation. We use the language model head of the monolingual model from EleutherAI/polyglot-ko-12.8b.
In the case of \textsc{Random-Init}, we randomly initialize with Gaussian distribution. 
\textsc{Multi-Init} indicates to leverage multilingual model representation by averaging its subword embedding as the main experiment.
The \textbf{Boldface} signifies the superior performances.
}
\label{tab:ablation_embedding}
\end{table*}

\begin{table*}[h]
\centering
\begin{adjustbox}{width=0.8\textwidth}
\begin{tabular}{cl|l|lll|ll}
\thickhline
\multicolumn{2}{c|}{}                             & \multicolumn{1}{c|}{\textbf{}}       & \multicolumn{3}{c|}{\textbf{Summarization (0-shot)}}                                                                              & \multicolumn{2}{c}{\textbf{Translation (3-shot)}}                                     \\ \hline
\multicolumn{2}{c|}{\textbf{Lang}}                & \multicolumn{1}{c|}{\textbf{Method}} & \multicolumn{1}{c|}{\textbf{ROUGE-2}} & \multicolumn{1}{c|}{\textbf{ROUGE-L}} & \multicolumn{1}{c|}{\textbf{Tokens/sec}} & \multicolumn{1}{c|}{\textbf{BLEU}} & \multicolumn{1}{c}{\textbf{Tokens/sec}} \\ \hline
\multicolumn{2}{c|}{\multirow{2}{*}{{\textsc{Ko}}}} & MuMo                                 & \multicolumn{1}{l|}{20.3}             & \multicolumn{1}{l|}{35.9}             & 55.3                                     & \multicolumn{1}{l|}{21.7}          & 50.9                                     \\ \cline{3-8} 
\multicolumn{2}{c|}{}                             & w/o Verification                     & \multicolumn{1}{l|}{11.0(-9.3)}       & \multicolumn{1}{l|}{26.4(-9.5)}       & 60.8(+5.5)                               & \multicolumn{1}{l|}{16.3(-5.4)}    & 62.3(+11.4)                              \\ \hline
\multicolumn{2}{c|}{\multirow{2}{*}{{\textsc{Ja}}}} & MuMo                                 & \multicolumn{1}{l|}{11.6}             & \multicolumn{1}{l|}{26.3}             & 59.2                                     & \multicolumn{1}{l|}{24.3}          & 58.3                                     \\ \cline{3-8} 
\multicolumn{2}{c|}{}                             & w/o Verification                     & \multicolumn{1}{l|}{6.7(-4.9)}        & \multicolumn{1}{l|}{20.4(-5.9)}       & 69.1(+9.9)                               & \multicolumn{1}{l|}{10.8(-13.5)}   & 73.6(+15.3)\\ 
\thickhline
\end{tabular}
\end{adjustbox}
\caption{\textbf{Ablation Study}. 
While the exclusion of the verification accelerates approximately 1.2 times in inference speed, it significantly compromises the quality of the generation.
}
\label{tab:ablation_verify}
\end{table*}

\subsection{Comparative Analysis of Fine-Tuning Strategies}
\label{analysis:lang_adapt}
In the section, we provide a comparative analysis of three distinct fine-tuning strategies for multilingual models. This analysis aims to highlight the advantages and disadvantages of each strategy, particularly in terms of dataset requirements.
and the number of parameters to train.

\subsubsection{Setup}

The two strategies compared in the analysis are:

1. \textbf{Vanilla Fine-tuning}: This strategy, which serves as a baseline, involves fine-tuning a standard multilingual model on a target monolingual instruction dataset (44M tokens) without any modifications to the pre-trained vocabulary.

2. \textbf{Vocabulary Expansion}: Inspired by prior work~\cite{chau2020parsing,cui2023efficient}, this strategy involves expanding the vocabulary of the pre-trained multilingual model and fine-tuning on the instruction dataset. This method, unlike MuMo, expands not only the LM head but also the token embedding in the input layer. 
Two implementations of this strategy are considered. The first involves pre-training on large-scale text corpora (60B tokens)\footnote{We use the off-the-shelf checkpoint from \href{https://huggingface.co/beomi/llama-2-koen-13b}{beomi/llama-2-koen-13b}} before fine-tuning on the instruction dataset. 
This strategy is marked with a dagger in Table~\ref{tab:vocab_expansion}. The second only undergoes the fine-tuning phase on the instruction dataset.

To account for the variability of token unit between the different strategies, we report the inference speed with the  morphemes per second \textbf{(Morphemes/sec)}, providing a standardized measurement.\footnote{\href{https://pypi.org/project/python-mecab-ko/}{python-mecab-ko}}
We only compare the baselines in Korean, because of the availability of model.
\subsubsection{Discussion}

Table~\ref{tab:vocab_expansion} reveals a consistent trend across both summarization and translation tasks. The vocabulary expansion strategies, which expand the dimension of both the token embeddings and LM head, exhibit significant increases in inference speed, but this is accompanied by a substantial decrease in the quality of the generated output when not trained on large-scale text corpora. This indicates that merely fine-tuning with an expanded vocabulary on a limited downstream dataset may not suffice to maintain high-quality text generation, as suggested by \cite{conneau2019unsupervised}. Furthermore, while vocabulary expansion with pre-training achieves notable speed improvements, it does not exhibit significant enhancements in generation quality. 

In contrast, our proposed method exhibits a modest increase in speed while also slightly improving BLEU scores relative to vanilla fine-tuning. The principal advantage of our method lies in its capacity to attain these results without necessitating vast monolingual text corpora. This approach not only reduces the number of parameters that need to be fine-tuned, making it more parameter-efficient but also lessens the dependency on large-scale data for pre-training, making it a more data-efficient solution.

\subsection{Initialization of Target Monolingual LM Head}
We investigate the impact of three different initialization strategies on the target monolingual LM head $g_{\text{mono}}$ in the Target Monolingual LM head.
The first strategy involves leveraging embeddings that correspond to the pre-trained representation of a targeted monolingual LM head, termed as \textbf{\textsc{Mono-init}}.
The second strategy is initializing the parameters with random value using Gaussian distribution \textbf{(\textsc{Random-init})}.
Lastly, we utilize the embeddings from the pre-trained multilingual LM head \textbf{(\textsc{Multi-init})}, as the main experiment.
This is achieved by averaging the output embeddings of the multilingual model.

Table~\ref{tab:ablation_embedding} shows that \textsc{Multi-init} achieves a ROUGE-L score of 36.3 and a BLEU score of 21.7, which are close to the 36.2 ROUGE-L and 20.9 BLEU scores of \textsc{Mono-init}. 
On the other hand, \textsc{Random-init} shows a decrease in performance, with a ROUGE-L score of 35.5 and a BLEU score of 17.2. 

The result demonstrates that the \textsc{Multi-init} approach is almost equally effective with \textsc{Mono-init}. 
This suggests that our framework can be utilized some languages that have an off-the-shelf vocabulary set but lack suitable pre-trained representations.

\subsection{Effectiveness of Verification Step}

We design an ablation study to investigate the role of the verification step in the inference process ( Sec.~\ref{sec:verify}).
To assess the impact of the verification step, we generated sequences without employing the verification step. 

From the results in Table~\ref{tab:ablation_verify}, conducted in both Korean and Japanese, we notice that the overall generation speed is approximately 1.2 times faster when the verification is excluded.
However, it is crucial to highlight that the exclusion of the verification step in the inference phase leads to a significant reduction in the generation quality. 
This is evident in the decrease in ROUGE-2, ROUGE-L, and BLEU scores for both languages when the verification module is not used, as shown in the table. 
This suggests that while the verification step may slightly slow down the generation process, it plays a vital role in preserving the model's generation capability.

\subsection{Comparative Study in Single-Task Training}
\label{sec::mt_bench}
In the experiment, our primary objective is to investigate whether the inherent capabilities of the instruction-tuned multilingual model, which handles a variety of tasks, could be compromised when trained exclusively on single tasks using either Vocabulary Expansion or MuMo.
Both methods introduce newly initialized parameters, raising concerns about potential impacts on the model's versatility.
To address these concerns, we separately trained the model on each task - Question Answering (QA)~\citep{lim2019korquad1,kurihara2022jglue} and Summarization~\citep{hasan2021xl} - and subsequently conducted a comparative analysis between Vocabulary Expansion and MuMo.

For evaluation, we utilize multiple-task datasets, specifically Korean\footnote{\hyperlink{https://github.com/42dot/42dot_LLM/blob/main/eval/benchmark_set_v2.csv}{Korean-MT-bench}} and Japanese\footnote{\hyperlink{https://huggingface.co/datasets/shi3z/MTbenchJapanese}{Japanese-MT-Bench}}, which consist solely of questions.
For the measurement, We adopt the single-answer grading setup from LLM-as-a-judge~\citep{zheng2023judging}. 
This involves presenting a question along with model-generated answers to GPT-4 (acting as the judge) for assessment. The answers are graded on a scale from 1 to 10.

\begin{figure}[t]
\begin{adjustbox}{width=0.48\textwidth}
\centerline{\includegraphics[width=0.49\textwidth]{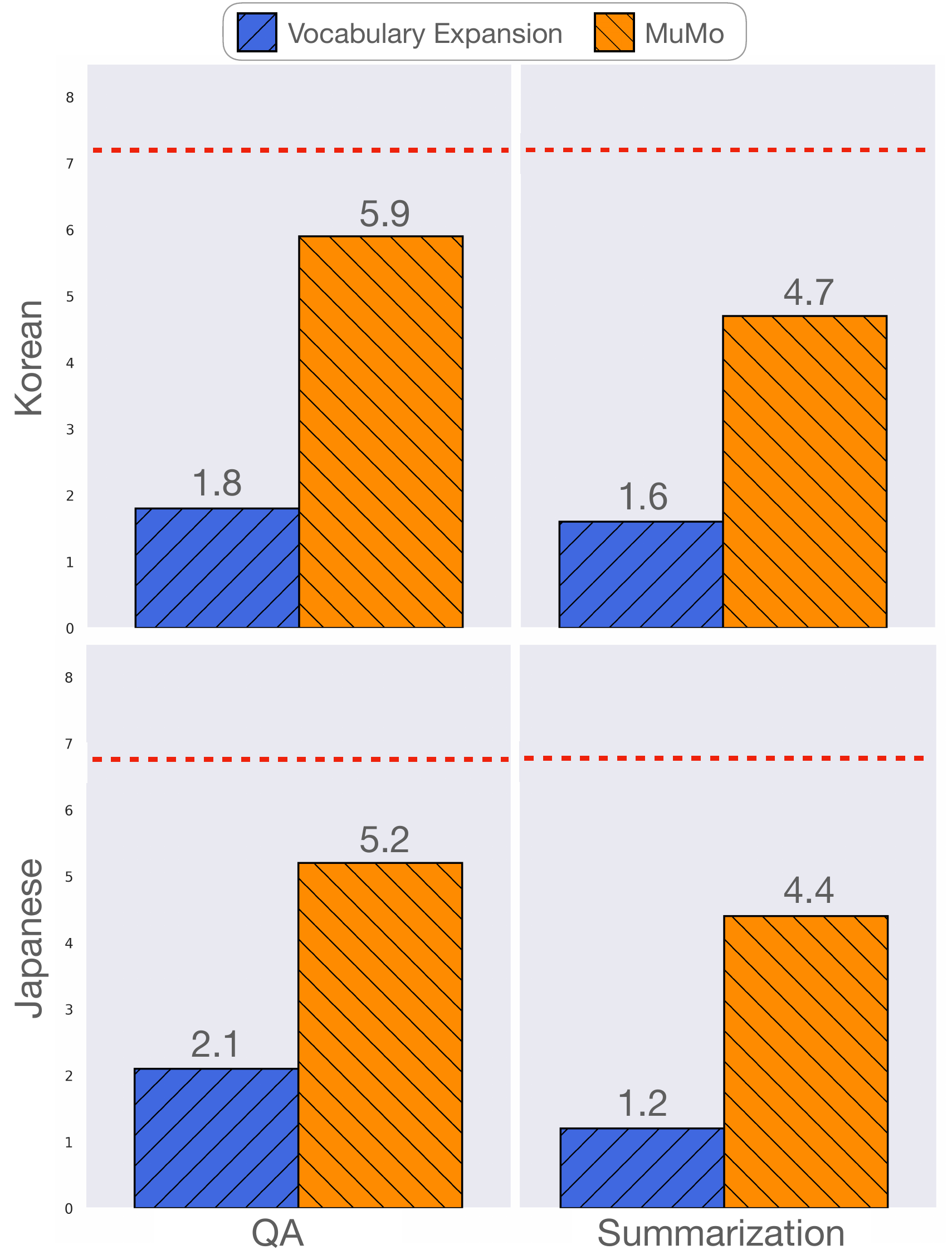}}
\end{adjustbox}
\caption{
\textbf{Evaluation on multiple-task after training on QA and Summarization task.}
The \textcolor{red}{red} dotted lines represent the average grading of single answers derived from the instruction-tuned multilingual language model. 
The decline is less pronounced with MuMo, suggesting its relative effectiveness in preserving the model's multi-task proficiency.
}
\label{fig:mt_bench}
\vspace{-0.4cm}
\end{figure}

As depicted in Figure~\ref{fig:mt_bench}, the instruction-tuned model initially achieves an average grading of 7.2 in the Korean experiment. 
However, when fine-tuned using only the QA task, Vocabulary Expansion receives a grading of 1.8, while MuMo receives a grading of 5.9. 
When trained solely on the summarization task, Vocabulary Expansion receives a grading of 1.6, while MuMo receives a grading of 4.7. 
Similar trends are observed in the Japanese experiment. 
The original model receives an average grading of 6.8. 
When fine-tuned with only the QA task, Vocabulary Expansion receives a grading of 2.1, while MuMo receives a grading of 5.2. 
When trained exclusively on the summarization task, Vocabulary Expansion receives a grading of 1.2, while MuMo receives a grading of 4.4.

These results suggest that while the grading of the model decreases when trained on single tasks using either method, the decrease is less pronounced with MuMo.
This indicates that MuMo is more effective at preserving the model's multi-task proficiency compared to Vocabulary Expansion. 
However, it is also clear that neither method can fully maintain the model's original instruction-following abilities on multiple tasks when trained solely on single tasks. 
These findings suggest that the instruction dataset, which the model was originally trained on, is crucial for preserving the pre-trained model's capabilities.


\section{Conclusion}\label{sec:conclusion}Our study has successfully tackled the challenges in generating text for non-alphabet languages, particularly those associated with excessive fragmentation issues.  
The approach not only speeds up text generation but also paves the way for more efficient multilingual language applications. 
Our future work will broaden our experimental scope to languages that were not sufficiently represented in the pre-trained multilingual language model.

\label{analysis:mt}

\section*{Limitations}
Our proposed framework has not been evaluated with languages that exhibit excessive fragmentation issues, such as Tamil, Hebrew, and Arabic~\citep{ahia2023all,petrov2023language}.
These languages were not explicitly mentioned in the pre-training corpus of Llama-2~\citep{touvron2023llama2}.
Additionally, our framework requires off-the-shelf tokenizers for target languages to make Target monolingual LM Head.
Our method does not alter the input sequence length, as we focus solely on improving the unit of prediction. This approach 
This approach differs from the the previous studies~\citep{rust-etal-2021-good,cui2023efficient} which efficiently encode text at the input-level sequence length for excessively tokenized languages.
Furthermore, the language models evaluated in the study are restricted to a maximum size of 13B.
Larger models, such as Llama-2 30B or 70B, were not implemented due to constraints on available computational resources.

\section*{Acknowledgements}
We thank Chaehun Park and Daeyoung Kim for their valuable discussions and feedback on the paper.

\bibliography{custom}
\bibliographystyle{acl_natbib}

\clearpage
\appendix

\section*{Appendix}\label{sec:appendix}


\section{Dataset Details}
\paragraph{Training Data}
Our study employed a multilingual instruction dataset from \citet{chen2023msift}, encompassing Korean and Japanese, for multilingual instruction tuning. 
Specifically, we  utilized ShareGPT and Alpaca-GPT4 for each respective language.
The dataset comprises 56k, 55k, and 168k examples for Korean, Japanese and English respectively.
To train MuMo LM head, we use ShareGPT and Alpaca-GPT4~\citep{chen2023msift} in Korean and Japanese for each language.

\paragraph{Evaluation Data}
In summarization task, we use validation and test split of XLSum~\citep{hasan2021xl}, which consist of 1100 examples. 
We found that more than half of the samples within the validation and test split surpassed the maximum sequence length of Llama-2. 
Consequently, we filtered out examples exceeding 1536 tokens. 
From the remaining examples, we randomly selected 300 for our experiments.

Regarding translation task, the dev-test set of FLoRes-200~\citep{goyal2022flores} is employed, consisting of 1012 parallel examples across both languages. 
We randomly use 3 examples as 3-shot prompts from training set for individual run. 

When evaluating multiple-task benchmark dataset~\ref{analysis:mt}, we exclude examples in coding and math categories.



\section{Additional Results}

\paragraph{Experiment on other Language Model}
Table~\ref{tab:other_model}, and Table~\ref{tab:other_model} present the comparative study in Llama-1 13B~\citep{touvron2023llama} and Mistral 7B~\cite{jiang2023mistral} respectively.

\paragraph{Generation Results}
Table~\ref{appendix:generated_xlsum} and Table~\ref{appendix:generated_flores} present generated texts in summarization and translation tasks.


\section{Environment Details}
All experiments are implemented using an A100-40GB GPU. 
The library versions utilized across all experiments include Python 3.9.10, Pytorch 2.1.0, and Transformers 4.34.0.

\section{Hyperparameter Details}\label{appendix:generation}

\begin{table}[h]
\centering
  \begin{adjustbox}{width=0.48\textwidth}
  \small
    \begin{tabular}{l|c }
    \thickhline
    \textbf{Hyperparameter} &\textbf{Value}  \\ [0.1ex] 
    \hline
    Learning rate & 2e-5 \\
    \hline
    Epoch & 3 \\    
    \hline
    Dropout & 0.1 \\        
    \hline  
    Tensor Type & bfloat16 \\
    \hline
    Batch size & 128 \\
    \hline  
    Optimizer & AdamW \\
    \hline
    Weight decay & 0.01 \\
    \hline
    Warmup ratio & 0.04 \\
    \hline
    Maximum sequence length & 2048 \\
    \hline
    Learning rate scheduler & cosine \\
    \thickhline
    \end{tabular}
  \end{adjustbox}
  \caption{Hyperparameters settings for multilingual instruction tuning. We follow the \hyperlink{https://github.com/lm-sys/FastChat/blob/main/scripts/train_vicuna_13b.sh}{script} from FastChat Library.
}
  \label{hyperparams_instruction}
\end{table}

\begin{table}[h]
\centering
  \begin{adjustbox}{width=0.48\textwidth}
  \small
    \begin{tabular}{l|c }
    \thickhline
    \textbf{Hyperparameter} &\textbf{Value}  \\ [0.1ex] 
    \hline
    Learning rate & 1e-3 \\
    \hline
    Epoch & 3 \\    
    \hline
    Dropout & 0.1 \\        
    \hline  
    Tensor Type & bfloat16 \\
    \hline  
    Batch size & 128 \\
    \hline  
    optimizer & 1.05 \\
    \hline
    Weight decay & AdamW \\
    \hline
    Warmup ratio & 0.04 \\
    \hline
    Maximum sequence length & 2048 \\
    \hline
    Learning rate scheduler & cosine \\
    \hline
    $d_{\text{ffn}}$ & 1280 \\
    \hline
    non-linearity function $q$ & SwiGLU \\
    \thickhline
    \end{tabular}
  \end{adjustbox}
  \caption{Hyperparameters settings for training MuMo framework.}
  \label{tab:hyperparams_mumo}
\end{table}

\begin{table}[h]
\centering
  \begin{adjustbox}{width=0.48\textwidth}
  \small
    \begin{tabular}{l|c }
    \thickhline
    \textbf{Hyperparameter} &\textbf{Value}  \\ [0.1ex] 
    \hline
    temperature & 0.1 \\
    \hline  
    sampling & True \\
    \hline  
    $p$ for top-$p$ sampling & 0.7 \\
    \hline 
    repetition penalty & 1.05 \\
    \hline
    exponential decay length penalty & (256, 1.03) \\
    \hline 
    max sequence length & 512 \\
    \hline 
    $k$ for top-$k$ sampling & 20 \\
    \hline 
    \end{tabular}
  \end{adjustbox}
  \caption{Hyperparameter settings for inference.}
  \label{tab:hyperparams}
\end{table}

\begin{table*}[t]
\small
	\begin{tabularx}{\textwidth}{c X}
\toprule
		\textbf{Task} & \textbf{Evaluation Prompt} \\
\toprule
		\textbf{Summarization}  
&
A chat between a curious human and an artificial intelligence assistant. The assistant gives helpful, detailed, and polite answers to the human's questions.

\# Document

\textcolor{blue}{\{\{}sourceDocument\textcolor{blue}{\}\}}

\#\# HUMAN: Summarize the document into a \textcolor{blue}{\{\{}targetLang\textcolor{blue}{\}\}} sentence.

\#\# ASSISTANT: 
\\ \hline
\textbf{Translation}  
& 
A chat between a curious human and an artificial intelligence assistant. The assistant gives helpful, detailed, and polite answers to the human's questions.

Translate the following text into \textcolor{blue}{\{\{}targetLang\textcolor{blue}{\}\}}.

\#\# HUMAN: \textcolor{blue}{\{\{}sourceString1\textcolor{blue}{\}\}}

\#\# ASSISTANT: \textcolor{blue}{\{\{}targetString1\textcolor{blue}{\}\}}

\#\# HUMAN: \textcolor{blue}{\{\{}sourceString2\textcolor{blue}{\}\}}

\#\# ASSISTANT: \textcolor{blue}{\{\{}targetString2\textcolor{blue}{\}\}}

\#\# HUMAN: \textcolor{blue}{\{\{}sourceString3\textcolor{blue}{\}\}}

\#\# ASSISTANT: \textcolor{blue}{\{\{}targetString3\textcolor{blue}{\}\}}

\#\# HUMAN: \textcolor{blue}{\{\{}sourceString\textcolor{blue}{\}\}}

\#\# ASSISTANT: 
\\
\toprule
\end{tabularx}
\caption{
The evaluation prompt for the main experiment (Sec.~\ref{sec:experiments}). 
We report on 0-shot results on summarization task, and 3-shot results on translation task respectively. 
}
\label{appendix:eval_prompt}
\end{table*}

\begin{table*}[t]
\small
	\begin{tabularx}{\textwidth}{c X}
\toprule
		\textbf{Task} & \textbf{Training Prompt} \\
\toprule
		\textbf{QA}  
&
A chat between a curious human and an artificial intelligence assistant. The assistant gives helpful, detailed, and polite answers to the human's questions.

\# Document

\textcolor{blue}{\{\{}context\textcolor{blue}{\}\}}

\#\# HUMAN: \textcolor{blue}{\{\{}question\textcolor{blue}{\}\}} 

\#\# ASSISTANT: \textcolor{blue}{\{\{}answer\textcolor{blue}{\}\}}
\\ \hline
\textbf{Summarization}  
& 
A chat between a curious human and an artificial intelligence assistant. The assistant gives helpful, detailed, and polite answers to the human's questions.

\# Document

\textcolor{blue}{\{\{}sourceDocument\textcolor{blue}{\}\}}

\#\# HUMAN: Summarize the document into a \textcolor{blue}{\{\{}targetLang\textcolor{blue}{\}\}} sentence.

\#\# ASSISTANT: \textcolor{blue}{\{\{}summary\textcolor{blue}{\}\}} 
\\
\toprule
\end{tabularx}
\caption{
The training prompt for the analysis on single-task prompt finetuning (Sec.~\ref{sec::mt_bench}). 
}
\label{appendix:eval_prompt}
\end{table*}

\begin{table*}[t]
\centering

\begin{adjustbox}{width=0.99\textwidth}
\begin{tabular}{c|c|cccc|ccc}
\thickhline
{}     & {}      & \multicolumn{4}{c|}{\textbf{Summarization (0-shot)}}                                                           & \multicolumn{3}{c}{\textbf{Translation (3-shot)}} 
\\ \hline
{\textbf{Lang}} & {\textbf{Method}} & 
\textbf{ROUGE-2} & \textbf{ROUGE-L} & \textbf{Tokens/sec} & \textbf{Speed Up} & 

\textbf{BLEU} & \textbf{Tokens/sec} & \textbf{Speed Up} 


\\ \hline
\multirow{2}{*}{\textsc{Ko}}        
& Vanilla Decoding                        & 14.7        &  31.2       &  29.0                           & 1.00x       & 18.6               & 29.7  
& 1.00x
\\

& MuMo (Ours)                        & 12.8        & 30.7       &   45.0   &  1.51x
& 18.1 & 43.0 & 1.49x
\\

\hline
\multirow{2}{*}{\textsc{Ja}}        
& Vanilla Decoding                         & 10.4        & 21.0        &  28.6               & 1.00x       
& 20.7              &  32.6
& 1.00x

\\ 
& MuMo (Ours)                        & 9.6        & 20.2       &   54.3   &  1.89x      & 20.0  & 53.8 & 1.64x 
\\
\hline

\end{tabular}
\end{adjustbox}
\caption{
\textbf{Comparative study of the inference speed in Llama-1 13B~\citep{touvron2023llama}.} 
The column labeled ``\textbf{Speed Up}'' represents the relative performance improvement in inference speed compared to the vanilla decoding method. 
}
\label{tab:other_model}

\end{table*}

\begin{table*}[t]
\centering

\begin{adjustbox}{width=0.99\textwidth}
\begin{tabular}{c|c|cccc|ccc}
\thickhline
{}     & {}      & \multicolumn{4}{c|}{\textbf{Summarization (0-shot)}}                                                           & \multicolumn{3}{c}{\textbf{Translation (3-shot)}} 
\\ \hline
{\textbf{Lang}} & {\textbf{Method}} & 
\textbf{ROUGE-2} & \textbf{ROUGE-L} & \textbf{Tokens/sec} & \textbf{Speed Up} & 

\textbf{BLEU} & \textbf{Tokens/sec} & \textbf{Speed Up} 


\\ \hline
\multirow{2}{*}{\textsc{Ko}}        
& Vanilla Decoding                        & 23.0        &  36.5       &  34.2                           & 1.00x       & 18.5               & 37.3  
& 1.00x
\\

& MuMo (Ours)                        & 22.8        & 36.9       &   56.4   &  1.65x
& 18.3 & 63.2 & 1.69x
\\

\hline
\multirow{2}{*}{\textsc{Ja}}        
& Vanilla Decoding                         & 12.9        & 26.4        &  33.5               & 1.00x       
& 27.2              &  36.8
& 1.00x

\\ 
& MuMo (Ours)                        & 13.2        & 26.3       &   64.8   &  1.93x      & 26.9  & 66.3 & 1.80x 
\\
\hline

\end{tabular}
\end{adjustbox}
\caption{
\textbf{Comparative study of the inference speed in Mistral 7B~\citep{jiang2023mistral}.} 
The column labeled ``\textbf{Speed Up}'' represents the relative performance improvement in inference speed compared to the vanilla decoding method. }

\label{tab:other_model}

\end{table*}


\begin{table*}[t]
\small
\begin{tabularx}{\textwidth}{c X c c}
\toprule
		\textbf{} & \textbf{Texts} (\textit{ko}) & \textbf{Tokens/sec} & \textbf{ROUGE-L} \\
\toprule[0.15ex]
		Document  & 환경부는 22일 사회관계장관회의에서 '1회용품 함께 줄이기 계획'을 추진한다고 발표했다. 2022년까지 일회용품 사용량을 35\% 이상 줄이는 것이 정부의 목표다. 종이 일회용 컵 사용 금지 현재 카페나 빵집 등에서 일회용 플라스틱 컵은 사용이 금지되지만, 종이컵은 사용이 가능했다. 하지만 2021년부터 종이컵 제공 또한 전면 금지된다. 식당, 카페, 급식소에서 플라스틱 빨대, 젓는 막대 등도 2022년부터 금지된다. 매장에서 머그잔에 음료를 받아 마시다 포장해서 가져가려는 경우에도 일회용 컵 사용에 따른 추가 비용을 내야 한다. 환경부는 '컵 보증금제' 재도입을 검토 중이다. 소비자가 커피 등 음료를 구매할 때 일정 금액의 보증금을 내고, 컵을 반환하면 그 돈을 돌려받는 방식이다. '컵 보증금제'는 과거 한 차례 도입됐다가 2008년 폐지됐다. 포장과 음식 배달에서 제공되는 일회용 식기류 무상 제공도 2021년부터 금지된다. 정부는 배달 음식 용기 또한 친환경 소재 또는 다회용기로 전환을 유도하겠다고 발표했다. 장례식장에서도 2021년부터 일회용 식기 용품 사용이 금지된다. 비닐봉지도 금지 현재 비닐봉지는 백화점이나 슈퍼마켓 등 대규모 점포에서는 사용이 금지되어 있다. 편의점 같은 종합 소매업이나 빵집 등에서는 유상으로 구매가 가능하다. 하지만 2022년부터는 제과점이나 가게에서도 일괄 금지된다. 호텔 등 숙박업소의 경우, 50실 이상의 시설에서는 2022년부터 샴푸, 린스, 칫솔 등 일회용 위생용품 무상 제공이 금지된다. 2024년부터는 모든 숙박업소에 일괄 적용된다. 택배 포장재 줄이기 최근 택배와 신선식품 배송이 급격히 늘면서, 환경부는 배송용 포장재 사용량 증가 해소를 위한 사업도 추진한다고 발표했다. 과대포장 문제가 제기된 배송 상품의 경우 포장기준을 강화하고, 업계와 협의해 종이 완충재와 테이프 없는 상자 등 친환경 포장재를 마련할 계획이다. 2020년부터 이미 포장된 제품을 다시 포장해서 묶어 판매하는 소위 이중 포장 행위가 금지된다. 올해 13살인 라니엘은 8살 때부터 강물에 떠내려온 쓰레기를 줍기 시작했다 다른 나라는 2018년 10월, 유럽연합은 바다 오염을 막기 위해 일련의 일회용 플라스틱 제품 사용을 완전히 금하는 법안을 통과시켰다. 유럽연합은 2021년부터 법안이 발효할 것으로 기대하고 있다. 금지 품목에는 플라스틱 식기류, 빨대, 면봉 등이 있으며 식품과 음료에 사용되는 플라스틱 컵 등 일회용 플라스틱 사용 역시 줄이도록 하는 내용을 담았다. 인도의 경우 2022년부터 일회용 플라스틱 사용이 전면 금지된다.  & & \\ \hline

        GT  & 2021년부터 카페에서 음료를 포장할 경우, 일회용 컵을 무상으로 사용하지 못한다. 
        &  &  \\ \hline

		{Vanila Decoding}  & 환경부는 2022년까지 일회용품 사용량을 35\% 이상 줄이는 것을 목표로 '일회용품 함께 줄이기 계획'을 추진한다고 발표했다. & 27.7 & 33.4 \\ \hline
  
		{MuMo}  & 환경부에 따르면 2022년까지 일회용품 사용량에서 35\% 이상 줄이기를 목표로 하며, 현재는 일회용 플라스틱 컵 사용이 금지되었으며 2021년부터는 종이컵 제공도 금지될 예정입니다.  & 47.2 & 38.1 \\
\toprule[0.15ex]
\end{tabularx}

\vspace{0.25cm}

\caption{Generated texts on summarization task in Korean.
The sample is extracted from the validation set of XLSum~\citep{hasan2021xl}. 
GT indicates the ground truth summary of the example.
}
\label{appendix:generated_xlsum}
\end{table*}

\begin{table*}[t]
\small
\begin{tabularx}{\textwidth}{c X c c}
\toprule
		\textbf{} & \textbf{Texts} (\textit{ja}) & \textbf{Tokens/sec} & \textbf{ROUGE-L} \\
\toprule[0.15ex]
		Document  & 
  \begin{CJK}{UTF8}{min}
  犬のマックスは16時間、女の子に寄り添った 女の子のオーロラちゃんは前の日から行方が分からなくなり、家族や警察など約100人が捜索に当たっていた。 クイーンズランド・サザンダウンズの自宅を出て原野に迷い込んだオーロラちゃんの後を、犬のマックスが追い、16時間近くずっと寄り添っていたとみられている。高齢のマックスは、目と耳が部分的に不自由。 1人と1匹が丘の斜面で一緒にいるのを、親族が21日朝に発見した。 オーロラちゃんの祖母、レイサ・マリー・ベネットさんは、自宅から約2キロ離れた場所で、オーロラちゃんの叫び声を聞いたと豪ABCに話した。 「大急ぎで山を駆け上がって上までたどりつくと、犬がこちらに向かってきて、オーロラのところへ一直線に連れて行ってくれた」 親類によると、気温が15度まで下がるなか、オーロラちゃんは犬のマックスと岩の下に避難していたという。 警察車両の横に立つマックス。動物の専門家によると、高齢な犬ほど人間とのつながりを特に重視するという クイーンズランド警察は、マックスの行動を称え、名誉警察犬の地位を与えた。 「3歳の子供なら、夜間とても怖かっただろうし、とても寒かったはずだ。犬が寄り添っていたおかげで、女の子は心強かっただろうし、寒くならなかったのだろう。明るい結末でよかった」とクレイグ・ベリー警部は話した。 ツイッターでは大勢が、マックスをほめちぎり、おやつをたくさんあげてほしいと書き込んだ。 なぜずっとそばに なぜマックスがずっとオーロラちゃんのそばを離れなかったのかについて、シドニー大学のポール・クリービー教授（動物行動学）は、高齢な犬ほど人間とのつながりを大事にするので、女の子の動揺を察知したのだろうと話す。 「もし女の子が泣いていたなら、犬は元気付ける行動をとった可能性が高い」とクリービー教授はBBCに話した。 「女の子のそばにずっといて、支えてあげるのが、なにより大事な行動だったはずだ」
  \end{CJK}
  \\ \hline
{GT}  & 
\begin{CJK}{UTF8}{min}
豪クイーンズランドの警察は21日午前、原野に迷い出て行方不明になった3歳少女を発見したと発表すると共に、家族の17歳になる牧牛犬が女の子に約16時間寄り添っていたと明らかにした。
\end{CJK} &  & \\ \hline

{Vanila Decoding}  & 
\begin{CJK}{UTF8}{min}
クイーンズランド州のサザンダウンズに住む小さな女の子オーロラが、家族や警察など約100人が捜索に当たっていたが、16時間前に行方不明になった。
\end{CJK}
 & 29.2 &27.8 \\ \hline
{MuMo}  & 
\begin{CJK}{UTF8}{min}
オーストラリアのクイーンズランド州サバーンダウンズ地域で行方不明になった子供を見つけ、オーストラリア警察の名誉警察犬になった高齢の犬マックス（Max）についてのジャーナル記事とともに、オーストラリア警察の名誉警察犬になったという内容です。
\end{CJK}
  & 57.3 &31.9  \\
\toprule[0.15ex]
\end{tabularx}
\caption{Generated texts on summarization task in Japanese. 
The sample is extracted from the validation set of XLSum~\citep{hasan2021xl}.
GT indicates the ground truth summary of the example.
}
\label{appendix:generated_xlsum}
\end{table*}

\begin{table*}[t]
\small

\begin{tabularx}{\textwidth}{c X c c}
\toprule
		\textbf{} & \textbf{Texts} (\textit{en}$\rightarrow$\textit{ko}) & \textbf{Tokens/sec} & \textbf{BLEU} \\
\toprule[0.15ex]
		Source  & 
   Since moving to the Catalan-capital, Vidal had played 49 games for the club.
  \\ \hline
GT  & 
바르셀로나로 이적한 후 비달은 클럽을 위해 49경기를 뛰었습니다.
  &  \\ \hline

Vanila Decoding  & 
바르셀로나로 이적한 이후로 비달은 이 클럽에서 49경기에 출전했습니다.
 & 27.2 & 22.8\\ \hline
MuMo  & 
바르셀로나로 이적했던 비달은 클럽에서 총 49경기를 출전했습니다.
  & 45.7 & 26.9 \\ 
\toprule[0.15ex]
\end{tabularx}

\vspace{0.25cm}

\begin{tabularx}{\textwidth}{c X c c}
\toprule
		\textbf{} & \textbf{Texts} (\textit{en}$\rightarrow$\textit{ko}) & \textbf{Tokens/sec} & \textbf{BLEU}\\
\toprule[0.15ex]
		Source  & 
Just after 11:00, protesters blocked traffic on the northbound carriage in Whitehall.\\ \hline
GT  & 11시가 막 지난 후, 시위대는 화이트홀에 있는 북쪽으로 향하는 마차들의 교통을 막았다.
\\ \hline
Vanila Decoding  & 
백알 화이트홀에서 오전 11시 15분경, 시위자들이 북쪽 차선을 차단하여 교통을 방해했습니다.
 & 27.6 & 5.1 \\ \hline
MuMo  & 
백알 화이트홀에서 오후 11시 이후, 시위대가 북쪽 선행 차량을 차단했습니다.
  & 44.3 & 4.3\\ 
\toprule[0.15ex]
\end{tabularx}

\vspace{0.25cm}

\begin{tabularx}{\textwidth}{c X c c}
\toprule
		\textbf{} & \textbf{Texts} (\textit{en}$\rightarrow$\textit{ja}) & \textbf{Tokens/sec} & \textbf{BLEU}\\
\toprule[0.15ex]
		Source  & 
      Since moving to the Catalan-capital, Vidal had played 49 games for the club.
  \\ \hline
{GT}  & 
\begin{CJK}{UTF8}{min}カタルーニャの州都に移って以来、ビダルはクラブで49試合に出場しました。\end{CJK}
 \\ \hline

{Vanila Decoding}  & 
\begin{CJK}{UTF8}{min}
バルセロナに移動してから、ビダルさんは約49試合でプレーしています。
\end{CJK}
 & 27.5 & 6.8 \\ \hline
{MuMo}  & 
\begin{CJK}{UTF8}{min}
バルセロナに移動してから、ビダルさんは約49試合でプレーしています。
\end{CJK}
  & 52.2 & 6.8 \\ 
\toprule[0.15ex]
\end{tabularx}

\vspace{0.25cm}

\begin{tabularx}{\textwidth}{c X c c}
\toprule
		\textbf{} & \textbf{Texts} (\textit{en}$\rightarrow$\textit{ja}) & \textbf{Tokens/sec} & \textbf{BLEU}\\
\toprule[0.15ex]
		Source  & 
   Just after 11:00, protesters blocked traffic on the northbound carriage in Whitehall.
  \\ \hline
{GT}  & 
\begin{CJK}{UTF8}{min}
11時すぎちょうどに抗議者たちはホワイトホールの北行き車両の交通を遮断しました。 \end{CJK} \\ \hline

{Vanila Decoding}  & 
\begin{CJK}{UTF8}{min}
11時過ぎに、ホワイトホールの北行線上で抗議者が交通を妨害しました。
\end{CJK}
 & 28.3 & 26.7\\ \hline
{MuMo}  & 
\begin{CJK}{UTF8}{min}
午前11時過ぎ、デモ隊はホワイトホールの北へ向かう馬車の交通を阻止した。
\end{CJK}
  & 50.3 & 25.0 \\ 
\toprule[0.15ex]
\end{tabularx}
\caption{Generated texts on translation task. 
The samples are extracted from the dev-test set of FLoRes-200~\citep{goyal2022flores}.
GT indicates the ground truth sentence of the example.
}
\label{appendix:generated_flores}
\end{table*}

\begin{table*}[t]
\centering
\begin{adjustbox}{width=0.7\textwidth}
\begin{tabular}{cl|cc|c}
\thickhline
\multicolumn{2}{c|}{}                       & \multicolumn{2}{c|}{\textbf{Summarization (0-shot)}}                 & \textbf{Translation (3-shot)}   \\ \hline
\multicolumn{2}{c|}{\textsc{\textbf{LM Head Initialization}}} & \multicolumn{1}{c|}{\textbf{ROUGE-2}}       & \textbf{ROUGE-L}       & \textbf{BLEU}          \\ \hline
\multicolumn{2}{c|}{\textsc{Mono-init}}              & \multicolumn{1}{c|}{{20.7}} & 36.2          & 21.5          \\ \hline
\multicolumn{2}{c|}{\textsc{Random-init}}            & \multicolumn{1}{c|}{19.2}          & 35.5          & 17.2          \\ \hline
\multicolumn{2}{c|}{\textsc{Multi-Init}}             & \multicolumn{1}{c|}{20.3}          & {36.3} & {21.7} \\ \hline
\multicolumn{2}{c|}{\textsc{Focus-Init}}             & \multicolumn{1}{c|}{20.8}          & {36.5} & {21.9} \\ 

\hline
\end{tabular}
\end{adjustbox}
\caption{\textbf{Comparative analysis for the initialization strategy }
We exploit FOCUS~\citep{dobler2023focus} embedding to initialize the Target monolingual LM Head. 
Our framework can be harmonically integrated with the initialization strategy of multilingual token embedding.
}
\label{tab:ablation_embedding}
\end{table*}

\end{document}